%% file: main.tex
\pdfoutput=1
\documentclass{ecai}

\usepackage{algorithm}
\usepackage{algpseudocode}
\usepackage{latexsym}
\usepackage{amssymb}
\usepackage{amsmath}
\usepackage{amsthm}
\usepackage{pifont}
\usepackage{booktabs}
\usepackage{enumitem}
\usepackage{graphicx}
\usepackage{color}
\usepackage{hyperref}       
\usepackage{url}            
\usepackage{booktabs}       
\usepackage{amsfonts}       
\usepackage{nicefrac}       
\usepackage{microtype}      
\usepackage{xcolor}         
\usepackage{subcaption} 
\usepackage{placeins}  
\usepackage{amsmath}
\usepackage{float}
\usepackage{caption}
\usepackage{titlesec} 
\usepackage{multirow}

\newcommand{\WINFlowNets}{WINFlowNets}
\newcommand{\shortcite}[1]{\cite{#1}}
\newcommand{\ex}{\mathbb{E}}
\newcommand{\BibTeX}{B\kern-.05em{\sc i\kern-.025em b}\kern-.08em\TeX}
\DeclareMathOperator*{\argmax}{argmax}

\begin{document}
\begin{frontmatter}
\paperid{123}
\title{WINFlowNets: Warm-up Integrated Networks Training of Generative Flow Networks for Robotics and Machine Fault Adaptation}

\author[A]{\fnms{Zahin}~\snm{Sufiyan}\thanks{Corresponding Author. Email: zsufiyan@ualberta.ca.}}
\author[A,B]{\fnms{Shadan}~\snm{Golestan}}
\author[C]{\fnms{Yoshihiro}~\snm{Mitsuka}}
\author[C]{\fnms{Shotaro}~\snm{Miwa}}
\author[A,B]{\fnms{Osmar}~\snm{Zaiane}}

\address[A]{University of Alberta}
\address[B]{Alberta Machine Intelligence Institute (Amii)}
\address[C]{Mitsubishi Electric Corporation}

\begin{abstract}
Generative Flow Networks for continuous scenarios (CFlowNets)
    have shown promise in solving sequential decision-making tasks
    by learning stochastic policies using a flow and a retrieval network.
    Despite their demonstrated efficiency
    compared to state-of-the-art Reinforcement Learning (RL) algorithms,
    their practical application in robotic control tasks is constrained
    by the reliance on pre-training the retrieval network.
    This dependency poses challenges in dynamic robotic environments,
    where pre-training data may not be readily available or representative of the current environment.
    This paper introduces WINFlowNets,
    a novel CFlowNets framework
    that enables the co-training of flow and retrieval networks.
    WINFlowNets begins with a warm-up phase for the retrieval network to bootstrap its policy,
    followed by a shared training architecture and a shared replay buffer for co-training both networks.
    Experiments in simulated robotic environments demonstrate that
    WINFlowNets surpasses CFlowNets and state-of-the-art RL algorithms
    in terms of average reward and training stability.
    Furthermore, WINFlowNets exhibits strong adaptive capability in fault environments, making it suitable for tasks that demand quick adaptation with limited sample data. These findings highlight WINFlowNets' potential for deployment in dynamic and malfunction-prone robotic systems, where traditional pre-training or sample inefficient data collection may be impractical.
\end{abstract}
\end{frontmatter}

\input{ecai2025/sections/1_introduction}

\input{ecai2025/sections/2_relatedWork}

\input{ecai2025/sections/4_method}

\input{ecai2025/sections/5_experiments}
\input{ecai2025/sections/6_discussion}

\input{ecai2025/sections/7_conclusion}

\bibliography{ecai25}
\end{document}

%% file: ecai2025/sections/1_introduction.tex
\section{Introduction}
\begin{figure}[t]
    \centering
    \begin{subfigure}{0.48\linewidth}
        \centering
        \includegraphics[width=\linewidth]{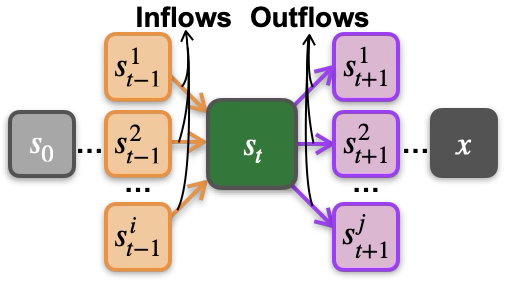}
        \caption{Directed acyclic graph representation of the decision-making process in GFlowNet.}
        \label{fig:intro_fig_a}
    \end{subfigure}
    \hfill
    \begin{subfigure}{0.48\linewidth}
        \centering
        \includegraphics[width=\linewidth]{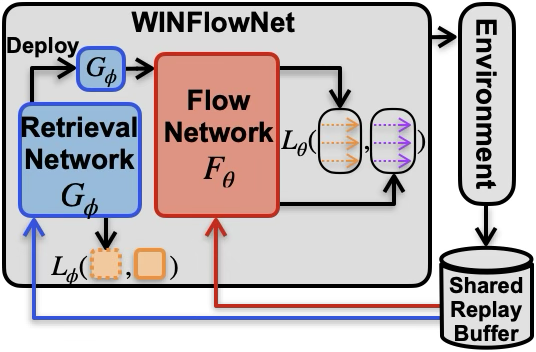}
        \caption{An overview of our proposed decision-making framework.}
        \label{fig:intro_fig_b}
    \end{subfigure}  
    \label{fig:intro_fig}
    \caption{Main concepts of \WINFlowNets.}
    
\end{figure}

Generative flow networks (GFlowNets)~\cite{bengio2021flow} 
is a class of generative algorithms
designed to learn a stochastic policy 
for generating (sampling) discrete structured objects 
(such as graphs, and sequences) $x \in \mathcal{X}$,
in the space of objects $\mathcal{X}$
to maximize a reward function $R(x)$.
To construct an object $x$,
GFlowNets starts from an initial null state $s_0$
and generates a trajectory by iteratively taking actions 
that add simple building blocks, 
passing through intermediate states
until it reaches the final state $x$ 
with a probability proportional to $R(x)$.
The process can be shown using directed acyclic graphs 
(depicted in Figure~\ref{fig:intro_fig_a}),
where nodes and edges denote the states and actions, respectively.
This problem can be framed as a sequential decision-making problem, 
making GFlowNets analogous to Reinforcement Learning (RL)~\cite{10.5555/3692070.3693625}.
GFlowNets learns a distribution over trajectories, 
enabling efficient exploration and reducing the risk of local optima.

GFlowNets has demonstrated potential 
in solving sequential decision-making problems
~\cite{NEURIPS2023_639a9a17,NEURIPS2024_2fb57276,pan2023stochastic,pan2023generative,jiralerspong2024expected}
and has been compared with RL algorithms across various problem settings, 
such as combinatorial optimization problems~\cite{jain2022biological,NEURIPS2023_27571b74},
HyperGrid modeling~\cite{bengio2021flow,niu2024gflownet}, and
operation scheduling~\cite{zhang2023robust}.
In several real-world domains, 
where the space of objects $\mathcal{X}$ is typically continuous,
a variation of GFlowNet, 
known as Continuous Flow Networks (CFlowNets)~\cite{li2023Cflownets} 
has been proposed.
CFlowNets generates a distribution over all possible trajectories,
and sample from it in proportion to the trajectories' performance.
As shown in various studies~\cite{li2023Cflownets,bengio2021flow,sufiyan2025studyefficacygenerativeflow},
GFlowNet/CFlowNets can be more sample-efficient than certain state-of-the-art RL algorithms such as Proximal Policy Optimiztion (PPO) 
and Soft Actor Critic (SAC).

Nevertheless, significant challenges remain in the training of CFlowNets~\cite{shen2023towards}.
While CFlowNets is effective for continuous spaces~\cite{pmlr-v202-lahlou23a}, 
its application can be impractical particularly in dynamic robotic control tasks~\cite{luo2024multi,sufiyan2025studyefficacygenerativeflow} 
such as object manipulation, where evolving environments require continual adaptation 
and where the space of objects $\mathcal{X}$ is exponentially large or composed of long trajectories~\cite{bengio2021flow,malkin2023gflownets}, 
making repeated re-training computationally expensive and inefficient.
To generate an object $x$, CFlowNets operates using two networks,
i.e., a flow network $F_{\theta}$, parametrized by $\theta$, and 
a retrieval network $G_{\phi}$, parametrized by $\phi$.
$F_{\theta}$ computes the flow, 
i.e., the probabilistic transitions 
between consecutive states conditioned on actions
(inflows and outflows in Figure~\ref{fig:intro_fig_a}).
This computation relies heavily on $G_{\phi}$, 
which estimates the possible parent states of a given state.
In the original CFlowNets training architecture,
$G_{\phi}$ is pre-trained prior to training $F_{\theta}$,
assuming a fixed data distribution.
However, this assumption limits the practical applicability of CFlowNets
in continuous robotic control tasks.


In robotic control tasks, 
environments often change due to evolving conditions and 
unexpected factors—particularly hardware faults 
and malfunctions~\cite{farid2022task}—resulting 
in out-of-distribution (OOD) scenarios. 
Without adaptation, 
robots are prone to errors in OOD situations, 
as they encounter previously unseen states more frequently. 
Such errors can slow down the learning process or 
gradually degrade the overall performance of CFlowNets.
This is especially problematic
because $G_{\phi}$ requires a separate pre-training dataset,
which may not be readily available in novel or dynamic environments.
If $G_{\phi}$ is pre-trained on data 
that does not fully represent the space of objects $\mathcal{X}$,
it can lead to both incorrect inflow and outflow estimations.
Moreover, pre-training $G_{\phi}$ introduces an extra step into the workflow, increasing resource requirements such as time, memory, and computational power.
Integrating the training of $G_{\phi}$ with $F_{\theta}$
is essential for enabling $G_{\phi}$ to dynamically adapt
to OOD situations.

This paper proposes \WINFlowNets{} (Figure~\ref{fig:intro_fig_b}), 
a novel CFlowNets framework that 
enables simultaneous training of 
both $G_{\phi}$ and $F_{\theta}$ networks
within a shared architecture.
To improve the training efficiency,
both networks utilize a shared reply buffer.
\WINFlowNets{} begins with a Warm-Up phase,
wherein $G_{\phi}$ independently interacts with the environment
to gather initial experience and refine its policy.
The Warm-Up phase is followed by a Dual-Training phase,
where both $G_{\phi}$ and $F_{\theta}$ actively interact with the environment,
collect experience, and contribute to the shared replay buffer
to support collaborative learning.
Experiments in simulated robotic environments demonstrate that 
\WINFlowNets{} outperforms standard CFlowNets 
in both training efficiency and adaptation performance 
under OOD situations. 
We also find that while \WINFlowNets{} surpasses state-of-the-art RL algorithms (PPO, SAC) in performance, 
its training efficiency remains comparatively lower.
Our contributions are: 
1) We propose \WINFlowNets, 
a novel training framework tailored for dynamic, 
continuous robotic control tasks.
2) We introduce and validate a two-phase training strategy—comprising a Warm-Up and 
Dual-Training phase—that enables efficient adaptation 
without the need for pre-training $G_\phi$.
3) We empirically demonstrate that 
\WINFlowNets{} outperforms standard CFlowNets 
and state-of-the-art RL algorithms 
in both standard and OOD (faulty) environments.
All code and data are publicly available at [link will be provided in the camera-ready version].

%% file: ecai2025/sections/2_relatedWork.tex
\section{Related Work}
\noindent\textbf{Reinforcement Learning (RL) for adaptation in robotic tasks}.
There is a rich body of literature proposing various RL 
~\cite{yang2021fault,chen2023adapt}
and Meta-RL
~\cite{ahmed2020complementary,luo2021mesa,nagabandi2019learning,nagabandi2019learningtoadapt,finn2017model}.
algorithms
for adapting to OOD situations in robotic tasks.
In a previous research, Meta-reinforcement learning (Meta-RL) frameworks have been developed to enable adaptation to unseen tasks using minimal data from task distributions. 
For instance, Neghabadi et al.~\shortcite{nagabandi2019learning}
proposed a model-based Meta-RL approach for fault adaptation tasks, utilizing two adaptive learners: a gradient-based adaptive learner (GrBAL) and a recurrence-based adaptive learner (ReBAL). GrBAL, based on model-agnostic meta-learning (MAML), uses neural networks to learn generalized dynamics, while ReBAL employs recurrent neural networks for task adaptation. These methods demonstrated significant sample efficiency, outperforming baselines like TRPO and MAML-RL, which required $1000$ times more data. Comparative analyses on continuous control tasks, such as Ant and HalfCheetah in OpenAI Gym, and real-life robots, showed faster adaptation to dynamic environments, underscoring the effectiveness of Meta-RL for fault adaptation.



\noindent\textbf{GFlowNets or CFlowNets for adaptation in robotic tasks}.
Few works have proposed flow networks for robotic control tasks~\cite{luo2024multi,brunswic2024theory,li2023Cflownets}. The closest work to ours is by 
Li et al.~\shortcite{li2023Cflownets}, where they applied CFlowNets to continuous control tasks like Point-Robot-Sparse, Reacher-Goal-Sparse, and Swimmer-Sparse, comparing them to RL algorithms such as PPO and SAC. In Point-Robot-Sparse and Reacher-Goal-Sparse, CFlowNets achieved higher average returns in fewer timesteps due to their superior exploration capabilities and ability to fit reward distributions. However, in Swimmer-Sparse, performance declined due to its steep reward gradient, which limited the exploration.
The study also highlighted that CFlowNets generated thousands of valid-distinctive trajectories, 
surpassing RL baselines, which struggled with exploration. RL algorithms like SAC and TD3 either generated fewer trajectories or saw performance drop as training progressed. These results underscore the strength of CFlowNets in solving sparse-reward tasks requiring diverse exploration but reveal limitations in tasks with dense reward structures.

Moreover, Sufiyan et al.~\shortcite{sufiyan2025studyefficacygenerativeflow} 
investigated the application of CFlowNets for robotic fault adaptation. 
Existing approaches often rely on RL, 
which can suffer from sample inefficiency and can struggle in OOD situations. 
The study introduced CFlowNets in the Reacher-v2 environment with four fault scenarios 
mimicking real-world robotic malfunctions. 
The authors compared CFlowNets with RL baselines, i.e., PPO, SAC, DDPG, and TD3 
highlighting their superior adaptation speed and asymptotic performance in most fault scenarios. 
We consider CFlowNets as a baseline in our study.

%

\noindent\textbf{Training Frameworks for Generative Flow Networks}.
Several related works have identified challenges in training GFlowNets and proposed suitable training frameworks to address them. For instance, Pan et al.\cite{pmlr-v202-pan23c} introduced a local credit assignment method that assigns partial rewards along a generative trajectory, enabling learning from incomplete trajectories and accelerating convergence. Similarly, Malkin et al.\cite{NEURIPS2022_27b51bac} proposed a global credit assignment approach that sums flow constraints over entire trajectories, rather than focusing solely on local transitions as in the standard training framework. This approach improves robustness to long action sequences and large action spaces. Finally, Shen et al.~\cite{shen2023towards} explored methods to enhance GFlowNet training efficiency and performance in limited-data settings. Specifically, they proposed an efficient sampling strategy for evaluation, combined with a relative edge flow policy parameterization to improve generalization to unseen states. However, the challenges of training GFlowNets/CFlowNets in dynamic robotic environments—where OOD situations, such as robotic faults or environmental changes, can disrupt the training process— has remained largely unexplored.

\noindent\textbf{Novelty of this work}.
We cast continuous robotic control tasks with OOD situations
as a generative flow networks problem. 
We then introduce a novel training architecture for CFlowNets,
called \WINFlowNets{},
and compare the resulting adaptation performance
with those found by CFlowNets, PPO, SAC, and DDPG.
We show that the combination of Warm-Up phase, Dual-Training phase,
and a share replay buffer for both $G_{\phi}$ and $F_{\theta}$
leads to improved performance and more effective adaptation in dynamic environments.

%% file: ecai2025/sections/4_method.tex
\begin{figure*}[t]
    \centering
    \includegraphics[width=0.8\textwidth]{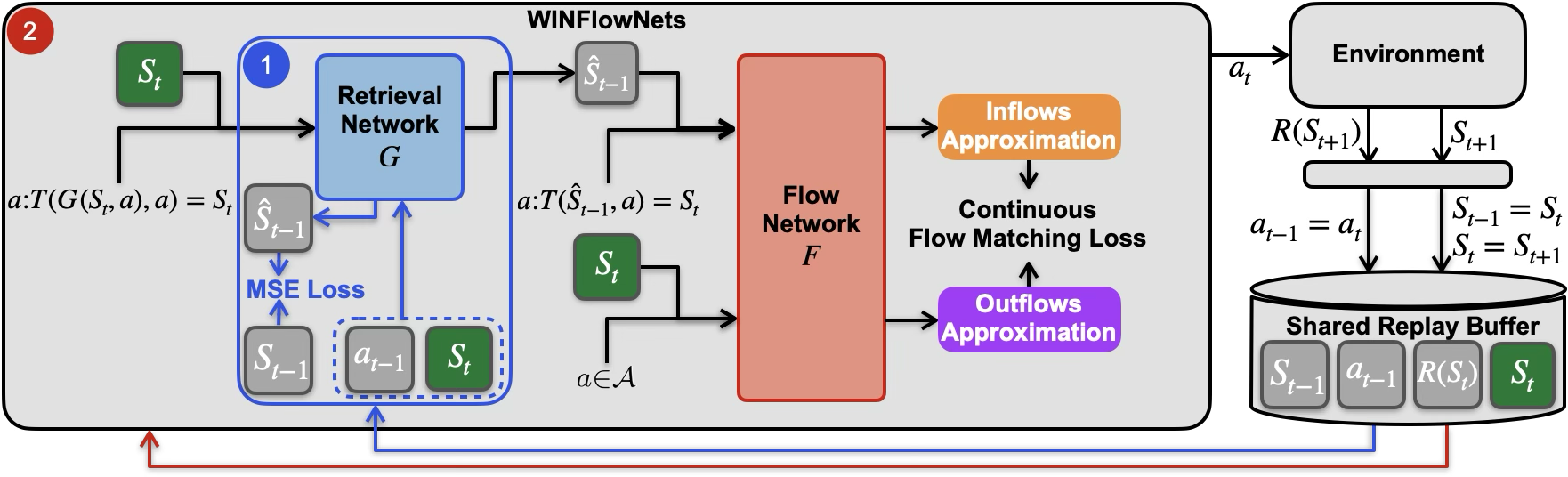}
    \caption{The architecture of \WINFlowNets.}
    \label{WINFlowNet}
\end{figure*}

\section{Method}
\subsection{Problem Setup}
Continuous robotic tasks are typically framed as
sequential decision-making problems,
where the environment is modeled  
as a Markov Decision Process (MDP),
denoted as $\mathcal{M}{=}(\mathcal{S}, \mathcal{A}, T, R, \gamma)$,
where $\mathcal{S}$ is the state space;
$\mathcal{A}$ is the action space;
$T$ is the environment's transition dynamics;
$R$ is the reward function; and
$\gamma{\in}[0,1]$ is the discount factor.
At each time $t{=}\{0,1,2,...,h\}$, with $h$ being the horizon, 
the agent interacts with the MDP via the allowed actions $\mathcal{A}$,  
and the state of the environment changes from 
$s_t{\in}\mathcal{S}$ to $s_{t+1}{\in}\mathcal{S}$ 
based on a transition dynamics $T(s_{t+1}|s_t, a_t)$, 
where $a_t{\in}\mathcal{A}$ is the action taken at time $t$.
Finally, the agent receives a feedback from the reward function $R(s_{t+1})$.
The objective of the agent is to find a policy $\pi$ 
to maximize an expected discounted sum of rewards 
$J(\pi){=}\ex_{\tau\sim\pi}[\sum_{t=0}^h {\gamma^t R(s_t)}]$ 
via a trial and error process, expressed as a trajectory
$\tau{=}(s_0{\rightarrow}s_1{\rightarrow}...{\rightarrow}x)$.
The agent's objective is to learn the optimal policy
$\pi^*{=}\argmax_\pi J(\pi)$.
We use $J(\pi)$ to compare the policies identified 
by our proposed method against those from baseline approaches.

\subsection{CFlowNets for Continuous Robotic Tasks}


The CFlowNets framework consists of three main components: 
action selection procedure,
flow matching approximation, and
continuous flow matching loss function.
In \textit{action selection procedure}, 
the agent interacts with the environment 
by iteratively sampling actions from a probability distribution $a_t{\sim}\pi(a_t|s_t)$.
Due to the continuous nature of the action space in continuous robotic tasks, 
the agent uniformly samples $M$ actions from the policy $\pi$ 
and generates an action probability buffer 
using the flow network $F_\theta$, defined as
$\mathcal{P}{=}[F_{\theta}(s_t, a_i)]_{i=1}^M$.
Actions with higher $F_{\theta}(s_t, a_i)$ values 
are sampled with higher probability.
The process repeats until a set of complete trajectories is constructed
and stored in a replay buffer.

The purpose of \textit{flow matching approximation} 
is to satisfy, for any state $s_t$, 
inflows ($f^+(s_t)$) and outflows ($f^-(s_t)$) are equal. 
This condition can be expressed as:
\begin{align} \label{flow_matching}
    f^+(s_t) &= f^-(s_t) \notag \\
    \int_{a:T(s,a)=s_t} F_{\theta}(s,a)\, da &= 
    \int_{a \in \mathcal{A}} F_{\theta}(s_t,a)\, da
\end{align}
These integrals are approximated by sampling $K$ actions 
from the action space $\mathcal{A}$.
To approximate inflows,
the retrieval network $G_\phi$ is used,
and then we calculate 
$\hat{f}^+{=}\frac{\mu(\mathcal{A})}{K}\sum_{k=1}^{K}{F_{\theta}(G_{\phi}(s_t, a_k), a_k)}$.
To approximate outflows,
we calculate $\hat{f}^-{=}\frac{\mu(\mathcal{A})}{K}\sum_{k=1}^{K}{F_{\theta}(s_t, a_k)}$,
where $\mu(A)$ indicates the measure of the continuous action space $\mathcal{A}$.
Finally, \textit{continuous flow matching loss function} can be derived by:
\begin{gather}
\mathcal{L_\theta} = \sum_{s_t = s_0}^{x} [\hat{f}^+ - \hat{f}^-]\notag\\
= \sum_{s_t = s_0}^{x} \Bigg[ \sum_{k = 1}^K F_\theta(G_\phi(s_t,a_k), a_k) - \lambda R(s_t) - \sum_{k = 1}^K F_\theta(s_t, a_k) \Bigg]^2 \notag \\
= \sum_{s_t=s_0}^{x} \Bigg[ 
\log\left(\epsilon + \sum_{k=1}^{K} \exp(F_{log\theta}(G_{log\phi}(s_t, a_k), a_k)) \right) \notag \\
- \log\left( \epsilon + \lambda R(s_t) + \sum_{k=1}^{K} \exp(F_{log\theta}(s_t, a_k)) \right) 
\Bigg]^2
\label{F_loss}
\end{gather}
where $\lambda$ is the scaling factor used to scale the summation of the sampled flows appropriately, considering the size of the continuous action space $\mathcal{A}$.
Note that in Equation~\ref{F_loss} it is assumed $G_\phi$ is pre-trained.
This assumption is often impractical or invalid 
in dynamic robotic environments,
particularly in OOD situations 
such as hardware faults.

\subsection{WINFlowNets Architecture}
We now introduce our proposed \WINFlowNets{} architecture.
The training mechanism of \WINFlowNets{} 
eliminates the requirement for pre-training the retrieval network $G_\phi$.
Figure~\ref{WINFlowNet} provides an overview of the \WINFlowNets{} architecture, 
including the training of the retrieval network $G_\phi$ 
(module 1 in Figure~\ref{WINFlowNet}), 
and the training of both the retrieval network $G_\phi$ 
and the flow network $F_\theta$ 
(module 2 in Figure~\ref{WINFlowNet}), and
its interaction with the environment at a given time $t$.
During the environment interaction, 
\WINFlowNets{} samples an action $a_t$ 
from an action probability distribution obtained 
by the forward propagation of \WINFlowNets{}.
Afterwards, \WINFlowNets{} interacts with the environment 
by taking action $a_t$,
which in turn updates the current state of the environment 
from $s_t$ to $s_{t+1}$,
and yields a reward of $R(s_{t+1})$.
After advancing to the next timestep ($t=t{+}1$),
we store a tuple of ${<}s_{t-1}, a_{t-1}, R(s_{t}), s_t{>}$
in a shared replay buffer, denoted as $\mathcal{B}$.


We modify the standard separated training loops of the $G_\phi$ and $F_\theta$ 
into a unified loop structure. 
In \WINFlowNets{}, both networks access the shared replay buffer $\mathcal{B}$, 
which contains state-action sequences that both networks encountered 
while interacting with the environment during the training process.
The shared replay buffer $\mathcal{B}$ 
facilitates the training of both the retrieval network $G_\phi$ 
and the flow network $F_\theta$
to estimate inflows and outflows of every state, respectively.


\noindent\textbf{Module 1: Warm-Up Phase of the Retrieval Network $\mathbf{G_\phi}$}.
\label{sec:G}
The Warm-Up phase is the foundational stage in the \WINFlowNets{}
for training $G_\phi$. 
The retrieval network $G_\phi$ is responsible for estimating
the predecessor states ($s_{t-1})$ given the current state $s_t$ and action $a_{t-1}$.
To train $G_\phi$, \WINFlowNets{} introduces a dedicated \emph{Warm-Up} phase
during which training of $G_\phi$ and $F_\theta$ 
remain active and inactive, respectively.
As described in Algorithm~\ref{alg:winflownets_warmup}, 
the agent begins each episode from an initial state $s_0$, and repeatedly 
(until reaching to a terminal state $x$ or time horizon $T$) 
samples a set of $M$ actions uniformly
from the action space $\mathcal{A}$.
An action $a_t$ is randomly selected from this set
and executed in the environment,
yielding the next state $s_{t+1}$ and
its corresponding reward $R(s_{t+1})$.
After advancing to the next timestep, 
The resulting tuple 
is stored in $\mathcal{B}$.

After each interaction episode, $G_\phi$ is trained
using a batch of $N$ tuples sampled from $\mathcal{B}$.
For each tuple $i$, $i\in\{1,...,N\}$,
we compute the predicted predecessor state as 
$\hat{s}_{t-1}^i{=}G_\phi(s_t^i,a_{t-1}^i)$,
and the prediction is compared to the ground-truth $s^{i}_{t-1}$
using Mean Squared Error (MSE) loss:
\begin{equation} \label{trainingG}
    \mathcal{L}_{\phi} = \frac{1}{N} \sum_{i=1}^N \left( \hat{s}_{t-1}^i - s_{t-1}^i \right)^2
\end{equation}
where $\hat{s}_{t-1}^i - s_{t-1}^i$ is the element-wise error
across the robot's state vector
(e.g., joint angles, velocities and fingertip-to-target position).
Standard gradient descent with the Adam optimizer is utilized 
to minimize Equation~\ref{trainingG}. 
Note that the shared replay buffer $\mathcal{B}$ 
continuously incorporates new data
during the training process, 
ensuring that $G_\phi$ can adapt to the dynamics of the environment.

Note that during the Warm-Up phase 
the learning rate $\eta$ 
begins at a low value and is gradually increased
to accelerate learning as training progresses.
This is because early in this phase,
$G_\phi$ operates with limited predictive accuracy and high uncertainty.
This conservative strategy 
ensures more stable updates on noisy gradient signals, 
while gradually increasing it enables faster convergence
as the agent gathers more diverse and representative data
from the environment.
We also define $w$ as the number of Warm-Up steps,
set to $100k$ based on our preliminary experiments,
providing sufficient time for $G_\phi$ 
to establish reliable predictions 
without unnecessary training overhead.


\noindent\textbf{Module 2: Dual Training of the Flow Network $\mathbf{F_\theta}$
and the Retrieval Network $\mathbf{G_\phi}$}.
\label{sec:F}
The flow network $F_\theta$ 
serves to approximate the inflows and outflows distributions
associated with a given state $s_t$,
a key step in maintaining the flow-matching condition expressed in Equation~\ref{flow_matching}.
After the Warm-Up phase,
during the Dual-Training phase,
both $F_\theta$ and $G_\phi$
are jointly optimized using data collected from agent-environment interactions.
As depicted in Algorithm~\ref{alg:winflownets_dualtraining},
each episode begins at an initial state $s_0$
and continues until reaching a terminal state $x$
or a time horizon $T$.
At each timestep,
the agent samples a set of candidate actions uniformly
from the action space $\mathcal{A}$,
and uses $F_\theta$ to construct 
the action probability buffer $\mathcal{P}$.
An action $a_t$ (selected based on $\mathcal{P}$),
is executed in the environment, 
and the resulting tuple 
is stored in $\mathcal{B}$.

After each episode,
a minibatch of transitions is sampled from $\mathcal{B}$.
For each state $s_t$ in the batch,
a set of parent state is predicted using $G_\phi$,
and these are used to compute the inflow approximation ($\hat{f}^+(s_t)$).
The outflow approximation ($\hat{f}^-(s_t)$) is computed directly using $F_\theta$
and the current state's candidate actions.
These inflow and outflow estimates are used to update $F_\theta$ via Equation~\ref{F_loss},
while both the ground-truth and estimated parent states 
are used to update $G_\phi$ through Equation~\ref{trainingG}.
The Dual-Training phase allows the networks to reinforce each other's learning, improving overall stability and performance in dynamic, OOD-prone environments.

\begin{algorithm}[tb]
\caption{WINFlowNets (Warm-Up phase)}
\label{alg:winflownets_warmup}
\begin{algorithmic}[1]
\State \textbf{Input:} $F_\theta$, $G_\phi$, $\beta$
\State \textbf{Initialize:} Warm-up duration $\omega = 100k$, dynamic learning rate $\eta$
\For{$t = 1$ \textbf{to} $\omega$}
    \State Set $s_t = s_0$
    \While{$s_t \neq$ $x$ \textbf{and} $t < T$}
        \State Uniformly sample $M$ actions $\{a_i\}_{i=1}^M$ from $\mathcal{A}$
        \State Select $a_t \sim\mathcal{U}(\{a_i\}_{i=1}^M)$
        \State Take $a_t$ and observe $(s_{t+1}, R(s_{t+1}))$
        \State $t \gets t + 1$
        \State Store $(s_{t-1}, a_{t-1}, R(s_t), s_t)$ in $\beta$
    \EndWhile
    \State Train $G_\phi$ using $\mathcal{L}_\phi = \frac{1}{N} \sum_{i=1}^{N} (\hat{s}_{t-1}^i - s_{t-1}^i)^2$
    \State $\eta = \eta + \epsilon$ \Comment{Increase $\eta$ to accelerate learning}
\EndFor
\end{algorithmic}
\end{algorithm}


\begin{algorithm}[tb]
\caption{WINFlowNets (Dual-Training phase)}
\label{alg:winflownets_dualtraining}
\begin{algorithmic}[1]
\While {not converged}:
    \State Set $s_t = s_0$
    \While{$s_t \neq x$ \textbf{and} $t < T$}
        \State Uniformly sample $M$ actions $\{a_i\}_{i=1}^M$ from $\mathcal{A}$
        \State Compute action probability buffer $\mathcal{P}$ via $\{F_\theta(s_t, a_i)\}_{i=1}^{M}$
        \State Sample action $a_t \sim \mathcal{P}$
        \State Take $a_t$ and observe $(s_{t+1}, R(s_{t+1}))$
        \State $t \gets t + 1$
        \State Store $(s_{t-1}, a_{t-1}, R(s_t), s_t)$ in $\beta$
    \EndWhile
    \State Sample minibatch $B$ from $\beta$
    \For{each state $s$ in $B$}
        \State Uniformly sample $K$ actions $\{a_i\}_{i=1}^K$ from $\mathcal{A}$
        \State Compute parent states $\{G_\phi(s, a_k)\}_{i=1}^K$
        \State Compute inflows:
        \Statex \hspace{3em} $\hat{f}^+=\log\left(\epsilon + \sum_{i=1}^K \exp\left(F_{\log\theta}(G_{\phi}(s_t, a_i), a_i)\right)\right)$
        \State Compute outflows:
        \Statex \hspace{3em} $\hat{f}^-=\log\left(\epsilon + \lambda R(s_t) + \sum_{k=1}^K \exp\left(F_{\log\theta}(s_t, a_k)\right)\right)$
    \EndFor
    \State Update $F_\theta$ via \(\mathcal{L}_\theta = \sum_{s_t = s_0}^{x} [f^+ - f^-] \)
    \State Update $G_\phi$ via \(\mathcal{L}_\phi = \frac{1}{N} \sum_{i=1}^{N} (\hat{s}_{t-1}^i - s_{t-1}^i)^2\)
\EndWhile
\end{algorithmic}
\end{algorithm}

%% file: ecai2025/sections/5_experiments.tex
\section{Experiments}
\subsection{Robot Control Setup}
\begin{figure}[tb]
    \centering
    \includegraphics[width=0.4\textwidth]{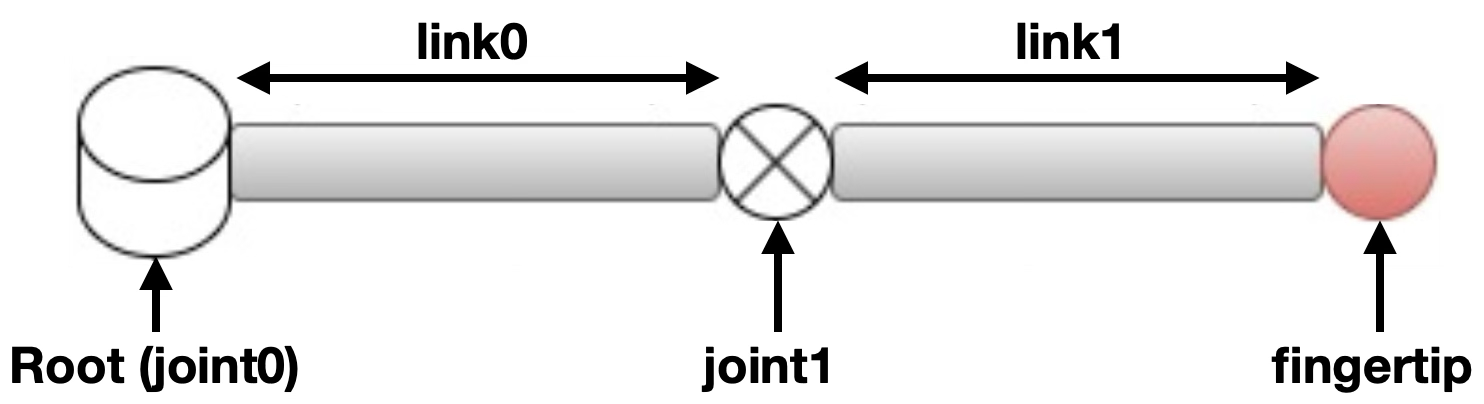}
    \caption{Visualization of the Reacher-v2 robotic arm.}
    \label{fig:mujoco}
\end{figure}

Similar to several related works
~\cite{li2023Cflownets,fournier2018accuracy, parisotto2019concurrent}, 
we opted for Reacher-V2\footnote{\url{https://gymnasium.farama.org/environments/mujoco/reacher/}}, 
a robotic simulation environment 
featuring a two-joint robotic arm 
with two degrees of freedom (Figure~\ref{fig:mujoco}). 
Each state in this environment is represented as a vector 
comprising joint angles, joint velocities, and the position of a target.
The target is a spatial point the robot must reach.
The two joints, named \texttt{joint0} and \texttt{joint1},
allow for a wide range of motion.
The first anchors the first segment of the arm (\texttt{link0}) to a fixed base (root),
and the latter connects the second segment (\texttt{link1}) to \texttt{link0}.
The robotic arm uses actuators to control its joints by applying torque.
Thus, the action space $\mathcal{A}$ is continuous 
and represented as a 2-dimensional vector of type \texttt{float32},
with each element ranging from $-1.0$ to $1.0$. 

The state space
includes the angles of the joints, 
the position of the end effector (fingertip), 
and the relative position to the target. 
We denote the state space $\mathcal{S}$ as \texttt{Box}(${-}\infty$, $\infty$, (11,), \texttt{float64}), representing an 11-dimensional continuous space where each element is a real-valued 64-bit floating-point number. 
The reward function in our task takes as input a state $s\in\mathcal{S}$ and an action $a\in\mathcal{A}$ and comprises two components: 
a distance penalty term, i.e., \(-\|p_{\text{fingertip}} - p_{\text{target}}\|\), 
where $p_{\text{fingertip}}$ and $p_{\text{target}}$ 
are the positions of the fingertip and the target, respectively, 
and a control penalty term, 
i.e., \(-\alpha\cdot \|a\|^2\) 
(with $\alpha$ ranging from $0$ to $1$),
which discourages excessive action magnitudes. 
Consequently, the reward function is given by:

\begin{equation}
    R(s, a) = -\|p_{\text{fingertip}} - p_{\text{target}}\| - \alpha \cdot \|a\|^2
\end{equation}


\begin{figure}[tb]
    \centering
    \includegraphics[width=0.45\textwidth]{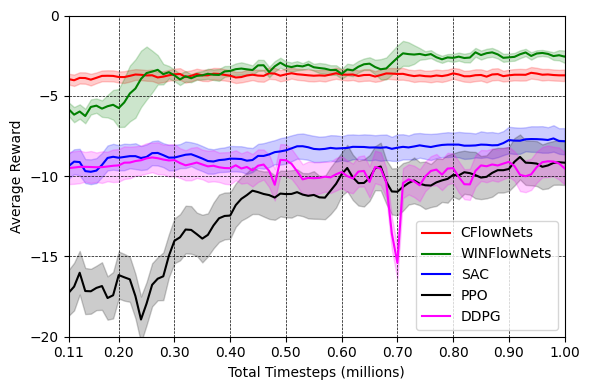}
    \caption{Average reward of CFlowNets, \WINFlowNets{}, and RL algorithms in the normal Reacher-v2 environment.}
    \label{fig:normal_env}
\end{figure}

\subsection{Normal Environment Case Study}
To establish a performance baseline,
we evaluate \WINFlowNets{} and all baselines
in the standard Reacher-V2 environment,
where no faults are introduced.
This setting serves as a controlled baseline
to evaluate the agent's performance in a stable and fully functional scenario.
We compare the performance of \WINFlowNets{} with CFlowNets and 
three standard RL algorithms (PPO, SAC, and DDPG), 
typically used as a benchmark for continuous robotic control tasks. 

\begin{table}[b]
\centering
\caption{Average final performance {$\pm$} standard deviation and sample efficiency (in millions of timesteps) of \WINFlowNets{} and baseline models in the normal Reacher-v2 environment. Bold entries indicate the highest-performing model.}
{%
\begin{tabular}{lcc}
\hline
\textbf{Model} & \textbf{Final Performance} & \textbf{Sample Efficiency} \\
\hline
SAC            & $-7.89{\pm}0.16$ & $0.67$ \\
PPO            & $-9.50{\pm}0.37$ & $3.39$\\
DDPG           & $-9.55{\pm}0.44$ & $5.20$\\
CFlowNets      & $-3.70{\pm}0.05$ & $\boldsymbol{0.10}$\\
\WINFlowNets{} & $\boldsymbol{-2.39{\pm}0.17}$ & $0.72$ \\
\hline
\end{tabular}%
}
\label{tab:normal_env_performance}
\end{table}

\subsection{Faulty Environment Case Study}
We evalute the adaptability of \WINFlowNets{} and all baselines
under OOD situations,
by introducing two common types of robotic faults:
\textit{Actuator Damage} (AD), and reduced \textit{Range Of Motion} (ROM).
AD can arise from overheating, gear failure, 
or electrical issues
~\cite{IndustrialAutomation2023servo, KEBAmerica2023}, 
which leads to weaker or unstable movements.
To simulate this, 
we reduce the actuator power (torque output)
to one-fourth of its original value.
This constraint affects the robot's ability 
to perform precise or forceful movements, 
resulting in degraded control.
ROM can arise from gear tear and wear,
mechanical restrictions,
or malfunctioning software
\cite{motioncontroltips2023, universalrobots2022}.
In Reacher-V2, we simulate this fault
by constraining the angular range of \textit{joint1} 
from its original $[-3.0, 3.0]$ radians to 
a narrower range of $[-1.5, 1.5]$.
This fault limits the robot's precision, flexibility
and overall maneuverability.
In this case study, 
we first train the models in the normal environment.
The learned model parameters, 
along with the replay buffer, are then transferred 
to resume training under the new fault condition.

\subsection{Results}
We collect empirical data in a span of 1 million timesteps.
Following related works (e.g.,~\cite{ijcai2022p477,sufiyan2025studyefficacygenerativeflow})
we evaluate the performance of the models in terms of
average reward, 
average final performance, and
sample efficiency
which are commonly used
to assess the performance of agents.
We calculate \emph{average reward}
($\pm$ standard deviation) over $10$ episodes,
showing the learning behavior of the models during 1 million timesteps.
To obtain \emph{average final performance}, 
we calculate both average and standard deviation of the average reward
over the last $20$ evaluations of 1 million timesteps.
This metric reflects the models behavior 
as training progresses towards the final steps, 
providing insight into the quality of the final policy learned.
We define \emph{sample efficiency}
as the timesteps required 
for each model to achieve their 
asymptotic performance, 
i.e., the average reward a model achieves once training has stabilized,
where stabilization is identified by monitoring the variance of rewards
across recent evaluations and observing no significant fluctuations\footnote{In our experiments, models reach stabilization within 5 million timesteps}.

\begin{figure}[tb]
    \centering
    \begin{subfigure}[b]{0.45\textwidth}
        \includegraphics[width=\textwidth]{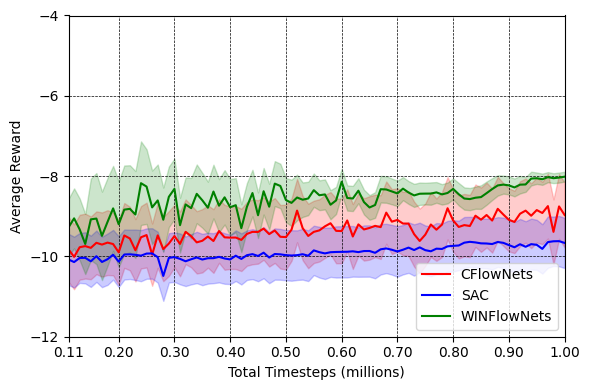}
        \caption{Actuator Damage (AD)}
        \label{fig:ad_env}
    \end{subfigure}
    \hfill
    \begin{subfigure}[b]{0.45\textwidth}
        \includegraphics[width=\textwidth]{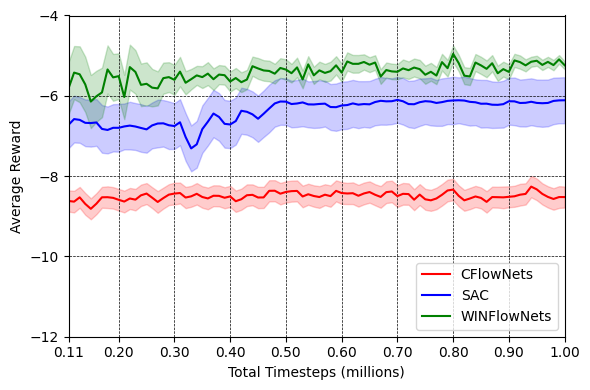}
        \caption{Reduced Range of Motion (ROM)}
        \label{fig:rom_env}
    \end{subfigure}
    \caption{Average reward of CFlowNets, \WINFlowNets{}, and RL algorithms in the Reacher-v2 environment under faulty conditions.}
    \label{fig:faulty_envs}
\end{figure}

\noindent\textbf{Normal environment}.
Figure~\ref{fig:normal_env}
illustrates the average reward of \WINFlowNets{} 
and our baselines across the normal environment
with shaded regions representing the $95\%$ 
confidence interval to indicate performance variability.
Note that the first evaluation begins at 110k timesteps,
following the 100k Warm-Up phase of \WINFlowNets{}.
According to Figure~\ref{fig:normal_env},
we make the following observations:
N1) \WINFlowNets{} outperforms CFlowNets and the RL baselines 
by a significant margin in the normal environment;
N2) \WINFlowNets{}, 
which does not rely on pre-training for $G_\phi$, 
shows a slower initial reward increase compared to CFlowNets.
However, it exhibits steady improvement and 
surpasses CFlowNets after approximately $240$k timesteps, 
ultimately achieving higher average reward by the end of training;
N3) CFlowNets exhibits a rapid rise in average reward, 
with the narrow-shaded region
highlighting the reliability and robustness 
provided by the pre-training of $G_\phi$;
N4) Among RL baselines, 
SAC follows a more stable learning curve,
and outperforms other RL models;
N5) DDPG and PPO shows unstable learning curves, 
with PPO in particular 
demonstrating an early performance dip
and requiring more timesteps (greater than 1 million) to stabilize.

To investigate these observations further,
we measure the average final performance
and sample efficiency of the models in Table~\ref{tab:normal_env_performance}.
It can be seen that \WINFlowNets{} shows
significantly better final performance, 
as confirmed by a two-tailed t-test ($p{<}0.05$).
However, 
it requires more samples than CFlowNets and SAC
to converge to its asymptotic performance.
This is primarily because in \WINFlowNets{}, 
both $G_\phi$ and $F_\theta$ 
co-adapt using a shared replay buffer, 
allowing them to continuously learn from the most recent interactions, 
and drive ongoing policy improvements.
For instance, note the two distinct surges in average reward 
around 0.2 and 0.8 million timesteps in Figure~\ref{fig:normal_env},
which highlight this adaptive behavior.

\noindent\textbf{Faulty environments}.
For faulty environments,
we choose SAC as our RL baseline, 
as it demonstrates the most stable performance
among the RL models.
In contrast, based on observation N5, 
PPO and DDPG are less suitable for comparison in faulty environments.
Figure~\ref{fig:faulty_envs}
illustrates the average reward of \WINFlowNets{} 
and our baselines across AD and reduced ROM environments.
Our observations in faulty environments are:
F1) \WINFlowNets{} outperforms CFlowNets and SAC;
F2) In both faulty environments,
\WINFlowNets{} shows unsteady performance
during the initial 400k timesteps
due to its more extensive exploration
enabled by the Warm-Up and Dual-Training phases 
(which are absent in CFlowNets),
allowing $G_\phi$ and $F_\theta$ to continually adapt
with new OOD data and ultimately find policies 
with higher average reward;
F3) \WINFlowNets{} initially shows high variance in performance,
comparable to SAC and CFlowNets,
but this variance progressively diminishes as training advances, 
indicating increased stability in learning.

We present the models' final performance and sample efficiency in Table~\ref{tab:faulty_env_performance}.
As shown, SAC and CFlowNeta are the most sample efficient models
in AD and reduced ROM environments, respectively.
This analysis complements observations F2 and F3
by highlighting the trade-off in \WINFlowNets{}:
although it requires more samples,
it ultimately achieves superior performance
via continual adaptation in OOD situations.
We believe the observed performance by \WINFlowNets{}
is because of the integration of the Warm-Up and Dual-Training phases, 
as well as the use of a shared replay buffer.
We investigate the effects of these phases in detail
in an ablation study.

\begin{table}[tb]
\centering
\caption{Average final performance {$\pm$} standard deviation and sample efficiency (in millions of timesteps) of \WINFlowNets{}, CFlowNets, and SAC in two faulty environments. Bold entries indicate the best-performing model in each setting.}
\resizebox{\columnwidth}{!}{%
\begin{tabular}{lccccc}
\hline
\multirow{2}{*}{\textbf{Model}} & \multicolumn{4}{c}{\textbf{Faulty Environment}} \\
\cline{2-5}
& \multicolumn{2}{c}{\textbf{AD}} & \multicolumn{2}{c}{\textbf{Reduced ROM}} \\
\cline{2-5}
& Final Perf. & Sample Eff. & Final Perf. & Sample Eff. \\
\hline
SAC            & $-9.69{\pm}0.19$ & $\boldsymbol{0.11}$ & $-6.16{\pm}0.03$ & 0.37 \\
CFlowNets      & $-9.01{\pm}0.17$ & 0.32 & $-8.50{\pm}0.08$ & $\boldsymbol{0.11}$ \\
\WINFlowNets{} & $\boldsymbol{-8.25{\pm}0.19}$ & 0.24 & $\boldsymbol{-5.25{\pm}0.12}$ & 0.12 \\
\hline
\end{tabular}
}
\label{tab:faulty_env_performance}
\end{table}

\subsection{Ablation Study} 
We evaluate the performance of CFlowNets and various 
\WINFlowNets{} implementations
on our normal environment. 
We examine three distinct versions of \WINFlowNets{}:
1) {\WINFlowNets{}-v1}
that has the Dual-Training phase without the Warm-Up phase and uses a shared replay buffer,
2) {\WINFlowNets{}-v2}
that incorporates both the Warm-Up and Dual-Training phases but allocates separate replay buffers for each network, and
3) the original \WINFlowNets{}
that has both the Warm-Up and Dual-Training phases with a shared replay buffer.
Note that we used this implementation in the previous section.


\begin{figure}[tb] 
    \centering
    \includegraphics[width=0.45\textwidth]{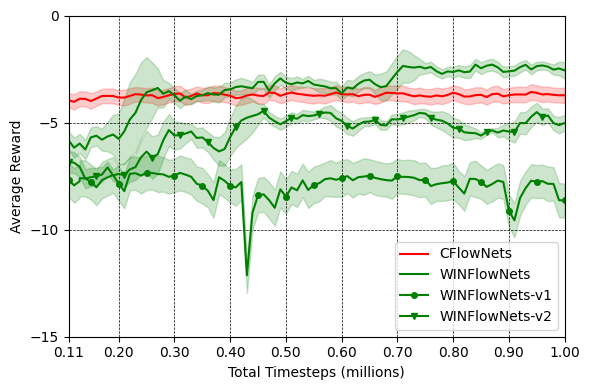} 
    \caption{Average reward for Vanilla CFlowNets and \WINFlowNets{} variants on our normal environment.}
    \label{fig:model_variants}
\end{figure}

Figure \ref{fig:model_variants} presents the average reward of 
the original \WINFlowNets{},
\WINFlowNets{}-v1,
\WINFlowNets{}-v2, and
CFlowNets.
Note that the asymptotic performance of 
\WINFlowNets{}-v1 and
\WINFlowNets{}-v2
is -8.2 and -4.6, respectively.
According to the figure,
\WINFlowNets{}-v1 exhibits the weakest performance,
in terms of average reward and confidence interval, 
indicating instability and higher variance during training.
While \WINFlowNets{}-v2 initially lags behind CFlowNets 
and performs comparably to \WINFlowNets{}-v1 
during the first 200k timesteps, 
its performance remains steady thereafter, with narrow confidence intervals indicating consistent and stable learning.
Nevertheless, the original version of \WINFlowNets{}
outperforms both \WINFlowNets{}-v1 and \WINFlowNets{}-v2
indicating the combined impact of the Warm-Up and Dual-Training phases, 
along with the benefit of a shared replay buffer.

%% file: ecai2025/sections/6_discussion.tex
\section{Discussion}

We have found that
\WINFlowNets{} and CFlowNets
outperform standard RL algorithms 
commonly used for continuous robotic control tasks,
i.e., SAC, PPO, and DDPG
in terms of average reward and sample efficiency,
resulting in finding significantly better policies.
The key modifications to the training architecture 
of the original CFlowNets,
i.e., Warm-Up and Dual-Training phases
along side of a shared replay buffer,
enhance its performance in the robotic tasks.

We argue that 
\WINFlowNets{} eliminates the need for pre-training $G_\phi$ 
and learns to adapt 
to the dynamic nature of tasks, 
such as OOD situations,
by jointly training $G_\phi$ and $F_\theta$,
enabling continual policy improvements based on recent shared
interactions with the environment.
It is worth mentioning that 
\WINFlowNets{} requires a substantial number of timesteps
to match the performance of CFlowNets.
This is because inflows and outflows approximations are not accurate,
as $G_\phi$ requires training on more experiences.
Although the initial training period showed a wider confidence interval region, it quickly stabilized as more improved versions of the retrieval network were deployed as the training continued. 
Note that the wider shaded region at the beginning of the post-Warm-Up phase reflects the adjustment period for the first few retrieval network instances.
Consequently,
in applications where safety of the policies are critical~\cite{xiong2024provably}, 
such as human-robot interaction scenarios~\cite{varley2024embodied},\WINFlowNets{} may not be feasible, 
as the initial performance of the model could pose risks.

\begin{figure}[tb]
    \centering
\includegraphics[width=0.45\textwidth]{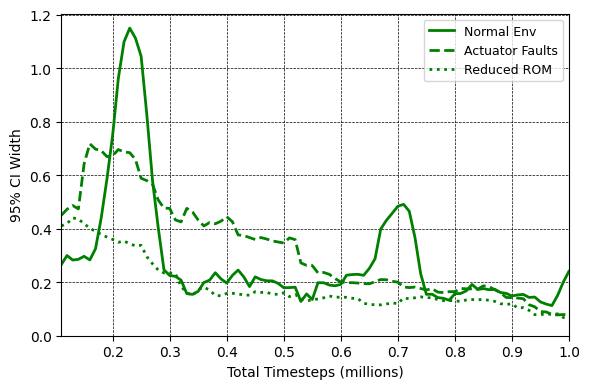}
    \caption{Confidence interval width of \WINFlowNets{} average reward over time in three environments.}
    \label{fig:uncertainty}
\end{figure}

To show this, we calculate in Figure~\ref{fig:uncertainty} 
the $\%95$ confidence interval width ($CI=2\times\frac{STD}{\sqrt{n}}$, where $n=10$ is the number of evaluation rollouts per point),
of \WINFlowNets{} average reward across training timesteps
in all environments, 
providing insights into how its performance uncertainty
evolves over time.
According to the figure,
the variance in performance is high initially
across all environments, 
especially in the normal environment.
However, the uncertainty decreases as training progresses, 
indicating stabilization and improved reliability of the learned policies.
Note that although the initial variance is high,
it remains comparable to those observed in our baselines.

The rapid improvement of CFlowNets 
is closely tied to their strong ability 
to learn effectively with fewer samples. 
This feature becomes apparent when contrasted with RL algorithms like PPO, 
which tend to struggle during early learning phases.
making them better suited for more deterministic environments,
where pre-training $G_\phi$ is feasible.

CFlowNets and \WINFlowNets{} consistently outperformed RL algorithms 
in both reward trends and sample efficiency on
all of the environments, except reduced ROM, where SAC outpeformed CFlowNets. 
A key reason for this superior performance is both \WINFlowNets{} and CFlowNets' ability to generate a distribution over trajectories, sampling paths with higher rewards more frequently. 
This distribution-based exploration reduces the risk of getting stuck in local optima, enabling \WINFlowNets{} and CFlowNets to converge toward globally optimal solutions. 
Also, CFlowNets' off-policy nature allows for efficient reuse of stored experiences, enabling faster convergence to asymptotic performance with minimal variance across runs.

In terms of RL algorithms, 
PPO's performance might be due to its on-policy nature, 
which relies on potentially suboptimal initial experiences. 
Additionally PPO's clipping mechanism, while providing stability, tends to limit the magnitude of policy updates, hence the slower learning process.
On the other hand, although SAC achieved lower average reward than \WINFlowNets{},
it demonstrated greater stability and superior sample efficiency relative to PPO and DDPG.
This can be due to its entropy-regularized exploration strategy, 
which prioritizes broader exploration of the action space
(better sample efficiency) 
over immediate reward maximization. 
Lastly, it is important to highlight that DDPG's 
sensitivity to hyperparameter configurations
and overestimation bias in its Q-value approximations 
leads to poor and inefficient learning.

\noindent\textbf{Limitations, future work, and potential applications}.
We wish to emphasize that \WINFlowNets{} demands 
significant computational overhead.
The shared replay buffer and Dual-Training phase 
contribute to increased GPU memory usage. 
As a result, the GPU consumption of \WINFlowNets{} 
exceeds that of its predecessor, CFlowNets, 
which was already noted to be resource-intensive in~\cite{sufiyan2025studyefficacygenerativeflow}. 
Consequently, \WINFlowNets{} is less suitable for systems that are highly resource-constrained.
In contrast, certain RL models may be more practical 
for deployment on resource-constrained platforms 
(for edge-device applications, see~\cite{yang2024beyond}), 
because of their relatively lower computational footprint at inference time.

Additionally, the performance of \WINFlowNets{} is highly sensitive 
to several hyperparameters, including the Warm-Up phase duration,
the learning rate of $G_\phi$, and the size of the shared replay buffer.
For instance, determining an appropriate the Warm-Up phase duration is essential,
as it can impact learning stability and performance. 
Note that longer durations may provide more robust initializations, 
reducing early-stage volatility, 
but at the cost of increased training time 
and potential overfitting to early-stage dynamics. 
In contrast, shorter durations may lead to faster adaptation 
but they risk instability or convergence to suboptimal policies.
The optimal duration is task-dependent and influenced by the complexity of the environment.
Consequently, further studies are required to 
systematically examine how task complexity affects the duration of the Warm-Up phase.
Nevertheless, the inclusion of the Warm-Up phase during training 
introduces latency which may not be ideal for continuous control task requiring fast convergence.
In such scenarios, SAC may be a more suitable alternative, 
offering faster early learning at the expense of suboptimal final performance.

While \WINFlowNets{} illustrates strong performance,
its sample efficiency can be improved.
This limitation is due to its Dual-Training phase,
which requires sufficient data for both networks.
One potential enhancement is to
incorporate ideas from prioritized shared replay buffer,
where experiences are sampled more frequently with higher learning potential,
such as
sampling those visited
experiences proportional to their absolute Temporal Difference (TD) errors. 
We refer readers to~\cite{pmlr-v180-pan22a} for more details.
Prioritizing the shared replay buffer can also shorten the duration of the Warm-Up phase.

We acknowledge that finding the optimal hyperparameter configuration 
requires extensive tuning, 
which may not be practical in dynamic environments. 
To address this,
we plan to reduce the computational overhead of \WINFlowNets{} 
by developing lightweight variants that
incorporate strategies such as model compression 
and distributed training for improved efficiency.

Lastly, another limitation of this study is that 
all experiments were conducted exclusively 
in a simulated robotic environment. 
As such, the performance observed may not fully generalize 
to real-world systems or environments with different dynamics.
future work will aim to extend the evaluation to broader set of environments and OOD situations,
including physical robotic platforms.

%% file: ecai2025/sections/7_conclusion.tex
\section{Conclusion}
This paper introduced \WINFlowNets{}, 
a novel framework for training CFlowNets in continuous control tasks. 
By integrating the training of the flow and retrieval networks 
via a shared replay buffer (Dual-Training phase) 
and incorporating a Warm-Up phase for the retrieval network.
We hypothesized that \WINFlowNets{} could enable more effective exploration of the action space, 
which contributes to identifying policies that yield higher average reward. 
To test this hypothesis, we
evaluated \WINFlowNets{} in three simulated Reacher-v2 robotic environments: 
one standard setting and two faulty conditions representing out-of-distribution scenarios.
We compared the performance of \WINFlowNets{} with CFlowNets, 
and state-of-the-art Reinforcement Learning (RL) algorithms, 
which are widely used for continuous control tasks.
Our experimental results demonstrate that 
\WINFlowNets{} outperforms CFlowNets and state-of-the-art RL algorithms (SAC, PPO, DDPG) 
in terms of average reward, while also finding policies with progressively lower uncertainty over time.
However, \WINFlowNets{}, like CFlowNets, remains more resource-intensive than RL algorithms, 
highlighting areas for future optimization.
